\documentclass[conference]{IEEEtran}
\IEEEoverridecommandlockouts
\usepackage{cite}
\usepackage{amsmath,amssymb,amsfonts}
\usepackage{algorithmic}
\usepackage{graphicx}
\usepackage{textcomp}
\usepackage{xcolor}
\usepackage{tabularx}
\usepackage{array}
\usepackage{hyperref}
\usepackage{xurl}
\usepackage{fancyhdr}
\usepackage{xcolor}
\usepackage{draftwatermark}
\SetWatermarkText{Author Version}
\SetWatermarkScale{0.35}
\SetWatermarkColor[gray]{0.95}

\def\BibTeX{{\rm B\kern-.05em{\sc i\kern-.025em b}\kern-.08em
    T\kern-.1667em\lower.7ex\hbox{E}\kern-.125emX}}
\begin{document}

\title{Benchmarking Local Language Models for\newline  
Social Robots using Edge Devices\\
\thanks{This work was in part supported by the Estonian Research Council grant (PRG3237).}
}

\author{\IEEEauthorblockN{Dorian Lamouille}
\IEEEauthorblockA{
\textit{University of Tartu}\\
Tartu, Estonia \\
\textit{ECAM LaSalle}\\
Lyon, France \\
}
\and
\IEEEauthorblockN{ Matevž B. Zorec}
\IEEEauthorblockA{\textit{Institute of Computer Science} \\
\textit{University of Tartu}\\
Tartu, Estonia \\
0009-0001-3334-9378}
\and
\IEEEauthorblockN{Farnaz Baksh}
\IEEEauthorblockA{\textit{Institute of Technology} \\
\textit{University of Tartu}\\
Tartu, Estonia \\
0009-0009-8362-7696}
\and
\IEEEauthorblockN{Karl Kruusam\"ae}
\IEEEauthorblockA{\textit{Institute of Technology} \\
\textit{University of Tartu}\\
Tartu, Estonia \\
0000-0002-1720-1509}
}

\maketitle
\thispagestyle{fancy}
\begin{abstract}
Social-educational robots designed for socially interactive pedagogical support, such as the Robot Study Companion (RSC), rely on responsive, privacy-preserving interaction despite severely limited compute. 
However, there is a gap in systematic benchmarking of language models for edge computing in pedagogical applications. 
This paper benchmarks 25 open-source language models for local deployment on edge hardware. 
We evaluate each model across three dimensions — inference efficiency (tokens per second, energy consumption), general knowledge (a six-category MMLU subset), and teaching effectiveness (LLM-rated pedagogical quality), validated against five independent human raters — using the Raspberry Pi~(RPi)~4 as the primary platform, with additional comparisons on the RPi~5 and a laptop GPU. 
Results reveal pronounced trade-offs: throughput and energy efficiency vary by over an order of magnitude across models, MMLU accuracy ranges from near-random to 57.2\%, and teaching effectiveness does not correlate monotonically with either metric. 
Among the evaluated models, Granite4 Tiny Hybrid (7B) achieves a strong overall balance, reaching 2.5 tokens per second, 0.90 tokens per joule, and 54.6\% MMLU accuracy; high MMLU accuracy does not appear necessary for strong teaching scores.
Human validation on four representative models preserved the automated rank ordering (Pearson $r = 0.967$, $n = 4$).
Based on these findings, we propose a three-tier local inference architecture for the RSC that balances responsiveness and accuracy on resource-constrained hardware. 
\end{abstract}

\begin{IEEEkeywords}
Benchmark, Social Robots, Language Model, Low-Compute Device, Educational Robots
\end{IEEEkeywords}

\section{Background}
Social robot companions powered by language models (LMs) can assist users through natural, conversational interaction, making them well-suited for educational support.
However, most deployments depend on cloud-based inference, raising concerns around cost, latency, and the privacy of sensitive user data \cite{kibriya_privacy_2024}. 
Local deployment on edge hardware \cite{open_source_LM} addresses these concerns but introduces severe memory and compute constraints, making model selection a critical design decision. 
Designing social-educational robots that rely on locally deployed language models requires balancing responsiveness, accuracy, energy consumption, and privacy. 
This trade-off is evident in practical open-source platforms such as the Robot Study Companion (RSC) \cite{baksh_open-source_2024}, which serves as a representative use case in this work.  
Accordingly, we benchmark open-source language models, measuring both hardware efficiency and teaching effectiveness. 

The main objective of this paper is to propose a reproducible benchmark for evaluating language models on edge devices in educational robotics contexts. 
In particular, the contributions of this paper include:
\begin{enumerate}
    \item \textbf{Open-source edge benchmark:} A reproducible evaluation framework measuring inference speed, energy efficiency, and pedagogical quality across 25 models on the Raspberry Pi (RPi)~4, with scalability comparisons on the RPi~5 and graphical processing units (GPU) (Section \ref{methods}).
    \item \textbf{Comparative model analysis:} Quantitatively comparing trade-offs relevant to model selection on resource-constrained educational robots (Section~\ref{sec:results}).
    \item \textbf{Preliminary caching architecture:} A three-tier local inference design that routes queries by semantic similarity, outlined for future implementation (Section~\ref{sec:discussion}).
\end{enumerate} 

\section{Related Works}

\subsection{Social Robots for Education}
Recent studies on social-educational robots explore integrations into higher education \cite{hri-highered}. 
However, there remains a need for versatile, low-cost, and accessible robotic platforms that can be deployed in resource-constrained educational and research environments.
Reproducible and open-source systems are particularly well-suited to these contexts, as they support transparency, customisation, and wider adoption.  

The RSC represents one such open-source platform designed to explore these requirements in practice \cite{baksh_open-source_2024}.
Its primary purpose is to provide interactive, multimodal learning support, thereby contributing to more engaging learning experiences.  
Initial studies report positive user acceptance, with students valuing the ease of use and perceived effectiveness \cite{baksh_university_2025}. 
The RSC project\footnote{\url{https://rsc.ee/}} charts a course toward more inclusive education by developing a secure, open robot platform that supports diverse learning styles and academic needs.

Adaptable robots can significantly enhance student engagement and academic performance. 
For instance, studies utilising an adaptive response-selection algorithm demonstrated that measures of task engagement were higher with adaptive robots compared to non-adaptive ones \cite{love2025adapting, donnermann_investigating_2022}.
Integrating Large LMs via a common application programming interface (API) represents one approach to enabling this adaptability to the RSC \cite{baksh_open-source_2024}.
However, it raises concerns, especially regarding privacy. 
The growing collection and potential misuse of personal data by large LMs have raised significant privacy concerns, necessitating the development of privacy-preserving techniques and local-language models\cite{kibriya_privacy_2024, chen2025survey}.

\subsection{Language Model for Education}

For a language model to function effectively as a teaching assistant, it must demonstrate broad and accurate knowledge across a wide range of educational subjects.
The Massive Multitask Language Understanding (MMLU) benchmark \cite{hendrycks_measuring_2021} is commonly used to evaluate such knowledge across diverse domains.
MMLU assesses model performance across 57 categories, ranging from secondary-level subjects to formal logic and applied ethics, using a total of 15{,}908 multiple-choice questions.
Each question presents four possible answers, focusing on factual correctness rather than token-level similarity.

MMLU was introduced to address limitations of earlier benchmarks such as GLUE \cite{wang_glue_2019}, which did not adequately capture model weaknesses across complex subject areas.
Random guessing yields an expected accuracy of 25\%, providing a clear baseline for interpretation.
Human domain experts achieve an average accuracy of approximately 89\% \cite{hendrycks_measuring_2021}, making MMLU a useful reference point for evaluating smaller and resource-constrained models.
For such models, achieving accuracy around 50\% represents a comparatively strong performance.
While MMLU effectively captures factual knowledge and instruction-following capability, it does not assess how explanations are structured or conveyed, which is particularly relevant in educational interaction.

EduBench addresses this limitation by evaluating pedagogical quality using human annotation across education-focused criteria \cite{xu_edubench_2026}.
Although EduBench provides richer insight into teaching behaviour, its reliance on human evaluation and server-class computation limits its applicability for benchmarking language models intended for deployment on resource-constrained edge devices.

\subsection{Language Model Inference on Edge Devices}
Several recent studies benchmark LM on resource-constrained hardware. 
Performance profiling of 25 quantised models across three single-board computers shows that these devices reliably support various models, though evaluations often prioritise system throughput and power over task-specific accuracy \cite{nguyen2025}.
Others investigate quantisation on the RPi \cite{ardakani_llmpi_2025}, reporting tokens per second, energy efficiency, and NUBIA response-quality scores, and mention a potential focus on social robots.

Similarly, runtime-level optimisations using LiteRT have demonstrated up to a 2x inference speedup over standard PyTorch frameworks with negligible impact on output quality \cite{yoon_litert_2025}.
The trade-offs between energy consumption and reasoning performance have been further detailed in a comprehensive study of 28 quantised models using hardware-based power measurements across five standardised datasets \cite{husom2025sustainable}.

These studies establish the technical feasibility of edge inference and, in the case of \cite{husom2025sustainable, nguyen2025}, chart the energy--accuracy frontier in depth. Yet none couple hardware efficiency with a task-grounded pedagogical evaluation, nor validate the automated evaluator against independent human raters. 
The present work closes both gaps: we jointly report hardware efficiency, MMLU, and LLM-rated teaching effectiveness on the same 25 models, then corroborate the teaching ranking against five human annotators on a representative four-model subset.

\section{Methodology} 
\label{methods}

\subsection{Evaluation Overview and Metrics}
\label{methods:evaluation}
To evaluate LMs for use on edge devices in educational settings, we evaluate each model along three complementary dimensions relevant to social-educational robots: (i) hardware efficiency under edge constraints, (ii) general knowledge and reasoning ability, and (iii) teaching effectiveness in educational interactions. 
Note: we estimate TPS from streamed output chunks rather than tokeniser-level token counts; Section~\ref{limitations} discusses the implications.

To assess general knowledge and reasoning ability, we used the MMLU benchmark \cite{hendrycks_measuring_2021}. 
Rather than executing the complete 57-category benchmark, we evaluate a representative six-category subset, including: Formal Logic, Global Facts, College Computer Science, College Mathematics, Marketing, and High School Macroeconomics. 
This subset comprises 1,050 test questions spanning reasoning-intensive and knowledge-intensive domains commonly encountered in secondary and undergraduate education, while remaining computationally tractable for comparative evaluation.
To conduct the MMLU evaluation, we used the DeepEval framework \cite{deepeval2024} (3-shot prompting; temperature~0.1). 
As MMLU measures model knowledge rather than inference performance, its accuracy is hardware-agnostic; execution infrastructure details appear in Section~\ref{methods:hardware-execution}.

\subsection{Teaching Effectiveness Evaluation}
\label{methods:teaching-effectiveness}

To assess pedagogical usability, we defined a fixed set of ten questions (Table \ref{tab:Benchmark Question}), derived from prior requirements for a robotic tutor, such as adaptable learning support \cite{baksh_open-source_2024}. 
The questions span explanatory depth, adaptability, misconception handling, and student guidance.
Each question targets a distinct aspect of pedagogical assistance.
Six items (Q1--Q4, Q6, Q7) use gravity as a controlled topic, isolating pedagogical competence from domain-difficulty variance while probing distinct functions from basic exposition to misconception handling; the remaining four items cover casual interaction, critical thinking, and student guidance.

All local models (see Table \ref{tab:model_benchmark}) generated responses to all questions in a single session. 

We report single-run values (see Sections \ref{methods:select-models}, \ref{methods:hardware-execution} for more details) and logged all responses (see our Zenodo repository \cite{lamouille_2026}).

Afterwards, GPT-4o-mini rated each answer on a 1--10 scale according to
eight teaching criteria: clarity, accuracy, engagement, structure, completeness,
appropriate level, use of examples, and actionability.
We apply the same criteria in the human-rater validation (\S\ref{results:human-validation})~\footnote{The rating prompt sits at \href{https://github.com/RobotStudyCompanion/Benchmark_LM/blob/v0.1/analyze_results.py\#L444}{\texttt{analyze\_results.py}, line~444} in v0.1.}.

\begin{quote}
\textit{[\ldots] Rate the following response on a scale of 1-10 based on its ability to teach effectively. [\ldots]
\begin{enumerate}
    \item \textbf{Clarity}: Is the explanation clear and easy to understand?
    \item \textbf{Accuracy}: Is the information correct?
    \item \textbf{Engagement}: Does it engage the learner?
    \item \textbf{Structure}: Is it well-organised?
    \item \textbf{Completeness}: Does it adequately address the question?
    \item \textbf{Appropriate Level}: Is the language suitable for the intended audience?
    \item \textbf{Examples/Analogies}: Are helpful examples provided?
    \item \textbf{Actionable}: Does it provide practical next steps or ways to apply the knowledge? [\ldots]
\end{enumerate}
}
\end{quote}

Small language models are known to be sensitive to prompt length and formatting \cite{sclar2024quantifying, he2024prompt}. 
They may echo the prompt structure or follow meta-guidance, leading to inflated apparent performance by rewarding instruction-following rather than genuine teaching ability. 
Accounting for this effect, we evaluated models in two distinct parameter-size groups under size-dependent prompting conditions:
\newline 
\begin{enumerate}
    \item Models with more than 1.4 billion (B; $10^9$) parameters: tested with a structured prompt that included task purpose and guidance.
    \item Models with less than 1.4B parameters: tested with a minimal prompt consisting only of the question text.
\end{enumerate}

We selected 1.4B as the boundary because it corresponds to a natural gap in our model set (Table~\ref{tab:model_benchmark}) and aligns with the parameter range below which prompt echoing was visibly prevalent in pilot runs.
This split reduces the prompt-following confound within each group. 
It does, however, mean that cross-boundary comparisons (e.g.\ a sub-1.4B model against a larger one on the same question) remain partly confounded by prompting regime and warrant cautious interpretation; within-boundary rankings carry the stronger signal. 

\begin{table}[t]
\centering
\small
\renewcommand{\arraystretch}{1.3}
\caption{Benchmark Questions}
\label{tab:Benchmark Question}
\begin{tabular}{|
    >{\centering\arraybackslash}p{0.4cm} |
    >{\raggedright\arraybackslash}p{5cm} |
    >{\raggedright\arraybackslash}p{2.4cm} |
}
\hline
\textbf{\#} & \textbf{Questions} & \textbf{Use} \\
\hline
1 & Describe gravity in simple terms for someone who doesn’t know anything about it. & Simple knowledge testing \\
\hline
2 & Describe gravity for someone who knows about it and wants to know more about it. & Testing of the adaptation to different levels of the user \\
\hline
3 & Find a good analogy to describe gravity to someone. & Adaptability \\
\hline
4 & Give a small quiz (3 questions) about gravity. & Capacity to provide good learning support \\
\hline
5 & Do you think I need a big jacket if the weather is 10 degrees? & Casual talking \\
\hline
6 & Give a common myth about gravity and explain why it is false. & Capacity to deal with confusion \\
\hline
7 & Explain why a plane can fly even with gravity in simple bullet points. & Capacity to simplify \\
\hline
8 & What are the steps to write a master's thesis? & Student guidance \\
\hline
9 & What are the pros and cons of nuclear energy? & Critical thinking \\
\hline
10 & I am struggling with C++, where should I start? & Student guidance \\
\hline
\end{tabular}
\end{table}

\subsection{Model Selection and Prompting Strategy}
\label{methods:select-models}

For privacy-minded, local deployment on limited compute, these criteria guided our model selection and prompting strategy (described in Section~\ref{methods:teaching-effectiveness}):
\begin{enumerate}
\item \textbf{Size diversity:} Models spanning 270M to 8B parameters to characterise the performance-capability trade-off across the feasible range for edge deployment.
\item \textbf{Architectural diversity:} Standard decoder-only transformers, hybrid Mamba-transformer architectures, and reasoning-oriented variants.
\item \textbf{Open weights:} Only models with publicly available weights, excluding proprietary API-based systems.
\item \textbf{Recency:} Models released since 2023.
\item \textbf{Framework compatibility:} Availability through the Ollama inference framework to ensure reproducibility. 
\end{enumerate}

Table \ref{tab:model_benchmark} presents the twenty-five models selected according to these criteria. 
We chose quantisation levels to balance memory footprint and inference quality, respecting the RPi~4's constraints.

\begin{table}[t]
\centering
\small
\renewcommand{\arraystretch}{1.15}
\caption{Benchmark results for all 25 models on RPi~4 and 24 of 25 on RPi~5. TPS reported as RPi~4\,/\,RPi~5; ``---'' denotes a model not benchmarked on the RPi~5. All models use Q4\_K\_M quantisation unless noted.$^{\dagger}$ Per-model TPJ and laptop GPU values are available in the supplementary record~\cite{lamouille_2026}.}
\label{tab:model_benchmark}
\begin{tabular}{|
    >{\raggedright\arraybackslash}p{4cm} |
    >{\centering\arraybackslash}p{1.4cm} |
    >{\centering\arraybackslash}p{0.9cm} |
    >{\centering\arraybackslash}p{0.7cm} |
}
\hline
\textbf{Model Name} & \textbf{TPS} & \textbf{MMLU} & \textbf{Teach} \\
 (Params, GB)&(RPi4/RPi5)& \textbf{(\%)} & \textbf{(/10)} \\
\hline
Gemma3 (0.27B, 0.3\,GB)$^{\dagger}$ & 8.7\,/\,18.6 & 20.2 & 5.8 \\
\textbf{Qwen3 (0.6B, 0.5\,GB)} & 5.7\,/\,14.2 & 42.3 & 7.1 \\
TinyLlama (1.1B, 0.6\,GB)$^{\dagger}$ & 4.8\,/\,14.3 & 21.4 & 5.4 \\
\textbf{Gemma3 (1B, 0.8\,GB)} & 3.9\,/\,9.4 & 24.0 & 8.5 \\
Llama 3.2 (1B, 1.3\,GB)$^{\dagger}$ & 2.6\,/\,6.0 & 19.0 & 7.4 \\
Granite4 (1B, 3.3\,GB)$^{\dagger}$ & 0.89\,/\,2.1 & 47.7 & 8.0 \\
Granite4-h (1B, 1.6\,GB)$^{\dagger}$ & 1.6\,/\,--- & 26.8 & 7.8 \\
Falcon3 (1B, 1.8\,GB)$^{\dagger}$ & 2.1\,/\,5.2 & 21.8 & 7.4 \\
DeepSeek-R1 (1.5B, 1.1\,GB) & 3.2\,/\,9.2 & 42.0 & 6.3 \\
Qwen3 (1.7B, 1.4\,GB) & 2.5\,/\,6.6 & 57.2 & 7.8 \\
Exaone Deep (2.4B, 1.6\,GB) & 2.0\,/\,5.0 & 27.4 & 7.7 \\
Cogito (3B, 2.2\,GB) & 1.5\,/\,3.8 & 29.3 & 7.1 \\
Falcon3 (3B, 2.0\,GB) & 1.4$^{\S}$\,{/}\,5.2 & 30.1 & 7.1 \\
Granite4 (3B, 2.1\,GB) & 1.4\,/\,3.8 & 54.3 & 7.0 \\
Granite4-h (3B, 1.9\,GB) & 1.3\,/\,3.3 & 51.7 & 8.0 \\
Llama 3.2 (3B, 2.0\,GB) & 1.5\,/\,4.3 & 23.9 & 7.7 \\
Phi3 (3.8B, 2.2\,GB)$^{\dagger}$ & 1.3\,/\,3.4 & 52.9 & 6.9 \\
Phi4-mr (3.8B, 3.2\,GB) & 1.0\,/\,2.8 & 35.1 & 7.2 \\
Gemma3 (4B, 3.3\,GB) & 1.2\,/\,3.1 & 51.0 & 7.6 \\
Nemotron-mini (4B, 2.7\,GB)$^{\ddagger}$ & 1.4\,/\,3.8 & 0.0 & 6.2 \\
Qwen3 (4B, 2.5\,GB) & 1.2\,/\,3.2 & 46.9 & 7.6 \\
Mistral (7B, 4.4\,GB) & 0.64\,/\,1.9 & 49.4 & 8.0 \\
\textbf{Granite4 Tiny Hybrid (7B, 4.2\,GB)} & 2.5\,/\,6.4 & 54.6 & 7.8 \\
DeepSeek-R1 (7B, 4.7\,GB) & 0.67\,/\,2.0 & 44.9 & 8.1 \\
Cogito (8B, 4.9\,GB) & 0.62\,/\,1.9 & 45.0 & 7.8 \\
\hline
\end{tabular}
\vspace{0.5em}

{\footnotesize $^{\dagger}$Non-default quantisation: Gemma3 0.27B (Q8), TinyLlama (Q4\_0), Llama~3.2 1B and Falcon3 1B (Q8\_0), Granite4 1B (BF16), Granite4-h 1B (Q8\_0), Phi3 (Q4\_0). Granite4 1B (BF16) is not directly size-comparable to Q4-quantised peers.
$^{\ddagger}$Format violations; excluded from MMLU aggregate statistics.
Bold rows indicate the three models selected for the Robot Study Companion deployment (see also yellow highlighting in Figure~\ref{fig:benchmark_overview}).
$^{\S}$Falcon3 3B on RPi~4 includes two degenerate responses
(zero- and one-character outputs at Q8 and Q10, respectively);
the reported value is the unfiltered ten-question mean.}

\end{table}

\subsection{Hardware and Execution Platform}
\label{methods:hardware-execution}

As mentioned in Section~\ref{methods:evaluation}, MMLU accuracy is independent of inference performance.
Therefore, we conducted the subset MMLU evaluation using services provided by the University of Tartu High-Performance Computing centre with each model inferencing on an NVIDIA Tesla A100 (40GB) GPU to reduce evaluation time to 2–3 minutes per model.
 
To document performance parameters, we inferenced all 25 models on the RPi~4 Model B (8\,GB RAM), running RPi Lite OS (64-bit; released 24 Nov 2025; kernel v6.12; Debian v12-bookworm); the RPi~5 used the same OS image.
Computational and memory constraints of this platform helped inform model selection criteria, described in Section~\ref{methods:select-models}. 
We executed all models using Ollama\cite{ollama2024} with streaming enabled.
Additionally, we benchmarked 24 of the 25 models on the RPi~5 (8\,GB RAM; granite4:1b-h omitted) and 23 of the 25 on a laptop equipped with an NVIDIA RTX 4060 GPU (granite4:1b-h and granite4:3b-h omitted), establishing scalability baselines across hardware relevant to social robotics deployments. 
From the RPi~4 results, we selected three models (qwen3:0.6b, gemma3:1b, and granite4:tiny-h) on joint throughput, MMLU, and on-disk footprint criteria as RSC candidates.
To cover all models within runtime budgets, we benchmarked across multiple sessions per platform and deduplicated into a single canonical dataset; within each (model, platform, question) triple, values remain single-run. Thermal drift may therefore introduce unquantified variance into absolute hardware metrics, while preserving relative rankings.
 
To estimate power draw on both RPi platforms, we used voltage readings from \texttt{vcgencmd} combined with linear CPU-load-based current estimation (idle: 0.6\,A; full load: 3\,A at 5\,V nominal, calibrated on the RPi~4). 
We compute Tokens Per Joule (TPJ) as the ratio of the total number of generated tokens and the estimated energy consumed during inference.
Given the linear current model, absolute TPJ values carry an estimated uncertainty of ±15–20\%, however, relative rankings remain stable.
Hence, this approach provides relative efficiency comparisons across models but should not be interpreted as a precise calorimetric measurement.
The accompanying supplementary record~\cite{lamouille_2026} documents per-query telemetry schemas for each platform.

\section{Results}
\label{sec:results}

We evaluated 25 open-source language models (Table~\ref{tab:model_benchmark}\footnote{Granite4-h denotes the Granite4 Hybrid variant, distinct from the larger Granite4 Tiny Hybrid (7B).}) on the RPi~4.
Results derive from consolidated, deduplicated single-run means aggregating per-query hardware measurements with model-level accuracy and teaching summaries \cite{lamouille_2026}.
Figure~\ref{fig:benchmark_overview} presents the full results; subsections below treat each dimension in turn.

\begin{table}[t]
\centering
\small
\renewcommand{\arraystretch}{1.15}
\caption{Per-category MMLU accuracy (\%) for models exceeding 40\% aggregate. Categories: Formal Logic (FL), Global Facts (GF), College Computer Science (CCS), College Mathematics (CM), Marketing (MK), High School Macroeconomics (HSM).}
\label{tab:mmlu_categories}
\begin{tabular}{|
    >{\raggedright\arraybackslash}p{1.9cm} |
    >{\centering\arraybackslash}p{0.5cm} |
    >{\centering\arraybackslash}p{0.5cm} |
    >{\centering\arraybackslash}p{0.5cm} |
    >{\centering\arraybackslash}p{0.5cm} |
    >{\centering\arraybackslash}p{0.5cm} |
    >{\centering\arraybackslash}p{0.6cm} |
    >{\centering\arraybackslash}p{0.5cm} |
}
\hline
\textbf{Model} & \textbf{FL} & \textbf{GF} & \textbf{CCS} & \textbf{CM} & \textbf{MK} & \textbf{HSM} & \textbf{Avg} \\
\hline
Qwen3 (1.7B) & 44 & 34 & 54 & 40 & 77 & 61 & 57.2 \\
Granite4 Tiny Hybrid (7B) & 50 & 31 & 46 & 39 & 71 & 58 & 54.6 \\
Granite4 (3B) & 44 & 26 & 37 & 41 & 81 & 57 & 54.3 \\
Phi3 (3.8B) & 48 & 32 & 38 & 35 & 62 & 63 & 52.9 \\
Granite4-h (3B) & 48 & 17 & 46 & 36 & 65 & 60 & 51.7 \\
Gemma3 (4B) & 42 & 26 & 40 & 34 & 78 & 51 & 51.0 \\
Mistral (7B) & 39 & 19 & 48 & 24 & 81 & 48 & 49.4 \\
Granite4 (1B) & 43 & 23 & 45 & 35 & 77 & 42 & 47.7 \\
Qwen3 (4B) & 34 & 27 & 45 & 46 & 64 & 47 & 46.9 \\
Cogito (8B) & 45 & 29 & 41 & 24 & 61 & 46 & 45.0 \\
DeepSeek-R1 (7B) & 33 & 34 & 41 & 32 & 50 & 53 & 44.9 \\
Qwen3 (0.6B) & 41 & 23 & 47 & 41 & 55 & 39 & 42.3 \\
DeepSeek-R1 (1.5B) & 48 & 24 & 44 & 40 & 42 & 45 & 42.0 \\
\hline
\end{tabular}
\end{table}

\subsection{Knowledge Performance (MMLU)}
\label{results:knowledge-perf}

A 6-task MMLU subset assessed knowledge performance (evaluation parameters as described in Section~\ref{methods:evaluation}). 
Model accuracy ranged from 19.0\% to 57.2\% (Table~\ref{tab:model_benchmark}). 
Ten out of twenty-five models perform near or below random baseline of 25\% (Fig. \ref{fig:benchmark_overview}[b]).
Qwen3 1.7B achieved the highest aggregate score, followed by the Granite4 family, which placed three models above 50\%.
As Table~\ref{tab:mmlu_categories} shows, Marketing consistently yielded the highest per-category scores across top-performing models, whereas Global Facts and College Mathematics proved most challenging.

\subsection{On-Device Inference Performance on the RPi~4}
\label{results:rpi4-perf}

Compact models dominated throughput and efficiency: Gemma3 270M led at 8.7 TPS and 3.3 TPJ, followed by Qwen3 0.6B and TinyLlama (Table~\ref{tab:model_benchmark}).
Granite4 1B warrants a separate note: its BF16 quantisation inflates the on-disk footprint to 3.3\,GB and constrains throughput to 0.89\,TPS, values atypical for a 1B-parameter model and not directly comparable to its Q4-quantised peers.
Among higher-accuracy models, Granite4 Tiny Hybrid stood out, maintaining mid-range throughput (2.5 TPS) and strong efficiency (0.90 TPJ) despite its 7B parameter count, owing to its hybrid Mamba-transformer mixture-of-experts architecture.

Measurement note: Reasoning-oriented models (DeepSeek-R1 1.5B and 7B, Phi4-mr 3.8B) emit internal chain-of-thought tokens in the runtime stream; their throughput and latency metrics may therefore appear optimistic relative to the useful output delivered.

\subsection{Teaching Effectiveness Evaluation}
\label{results:teaching-evaluation}

Teaching effectiveness (LLM-rated, 1–10) ranges from 5.44 to 8.5 across models. 
Gemma3 1B achieved the highest mean teaching score (8.5/10) despite near-random MMLU accuracy, whilst the fastest models (TinyLlama, Gemma3 270M) scored lowest (Table~\ref{tab:model_benchmark}).
Most models clustered between 7.0 and 8.1, suggesting a relatively narrow spread once a minimum capability threshold is met.
We present these results as exploratory rather than definitive, given the small sample size; Section~\ref{results:human-validation} reports a multi-rater validation study, and Section~\ref{sec:discussion} revisits the relationship between teaching scores and the other evaluation dimensions.

\begin{figure*}[!ht]
\centering
\includegraphics[width=\textwidth]{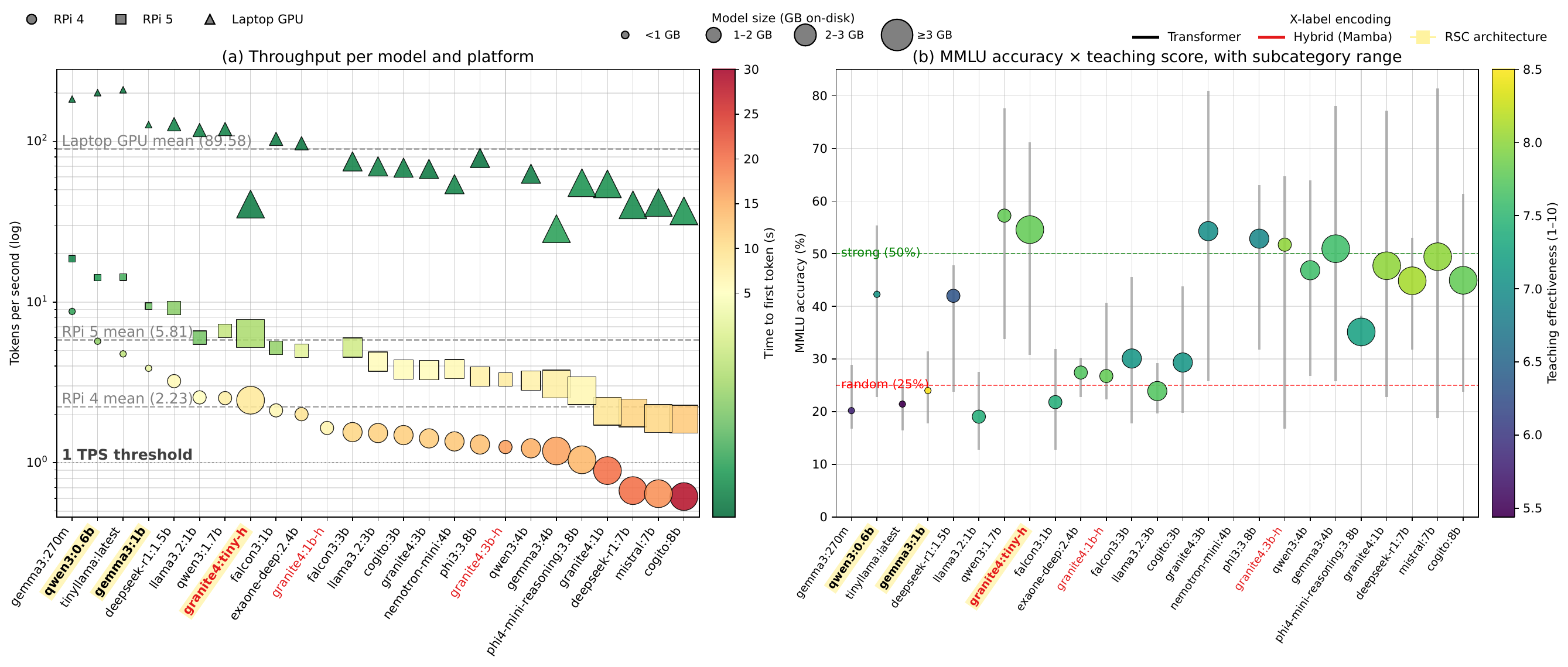}
\caption{Throughput scales by roughly an order of magnitude from Raspberry Pi~4 to laptop GPU, while MMLU accuracy and teaching effectiveness show no monotonic scaling with throughput (n=25 models on RPi~4, 24 on RPi~5, 23 on laptop GPU). 
Both panels share the x-axis, with models sorted by RPi 4 tokens per second (TPS) in descending order. 
X-axis labels are colour-coded by architecture (black: transformer; red: hybrid Mamba-transformer), with the three models selected for the Robot Study Companion (qwen3:0.6b, gemma3:1b, granite4:tiny-h) additionally highlighted in yellow. 
Marker shape encodes platform (circle: RPi 4, square: RPi 5, triangle: laptop GPU); marker size encodes the model's on-disk footprint in four tiers. 
(a) Throughput per model and platform, log scale. 
Marker hue encodes time-to-first-token (TTFT, seconds, diverging palette centred at 5s). 
Dashed grey lines mark per-platform mean TPS; the dotted line at 1 TPS indicates a usability threshold below which perceived responsiveness degrades noticeably. 
(b) Six-category MMLU accuracy (nemotron-mini:4b excluded per format-violation caveat). 
Vertical grey bars show the min–max range across the six subcategories. Marker hue encodes LLM-rated teaching effectiveness (1–10 scale, sequential viridis). 
Dashed red and green lines mark random-baseline (25\%) and strong-performance (50\%) reference points respectively.
}
\label{fig:benchmark_overview}
\end{figure*}

\subsection{Multi-Metric Trade-offs}
\label{results:metric-tradeoffs}

Jointly considering knowledge (MMLU), throughput (TPS), and energy efficiency (TPJ) shows that no single model maximises all dimensions. 
For example, Gemma3 270M leads on responsiveness but ranks near the bottom on MMLU, whereas Qwen3 1.7B achieves the highest MMLU at the cost of moderate throughput.
Granite4 Tiny Hybrid occupies a balanced intermediate position across knowledge, throughput, and efficiency dimensions (Table~\ref{tab:model_benchmark}).
Figure~\ref{fig:benchmark_overview}(b) visualises these trade-offs.

\subsection{Human Validation of Teaching Scores}
\label{results:human-validation}

To validate the automated GPT-4o-mini ratings, five independent raters scored a subset of 40~responses (4~models $\times$ 10~questions) against the same eight teaching criteria (section~\ref{methods:teaching-effectiveness}). 
Each rater evaluated all 40~items, yielding 200~annotated ratings in total.
All rater spreadsheets and the analysis notebook are available in the supplementary repository~\cite{lamouille_2026}.

Table~\ref{tab:human_vs_gpt} reports mean human scores alongside GPT-4o-mini scores per model. 
Human raters ranked Gemma3~1B highest ($M = 7.74$), followed by Granite4 Tiny Hybrid~7B (7.04), Mistral~7B (6.81), and Gemma3~0.27B (5.54). 
This ordering matched the automated evaluation, producing a Pearson correlation of $r = 0.967$ ($p = 0.033$) and a mean absolute difference of 0.75~points, with GPT-4o-mini scoring consistently higher.

\begin{figure*}[t]
\centering
\includegraphics[width=\textwidth]{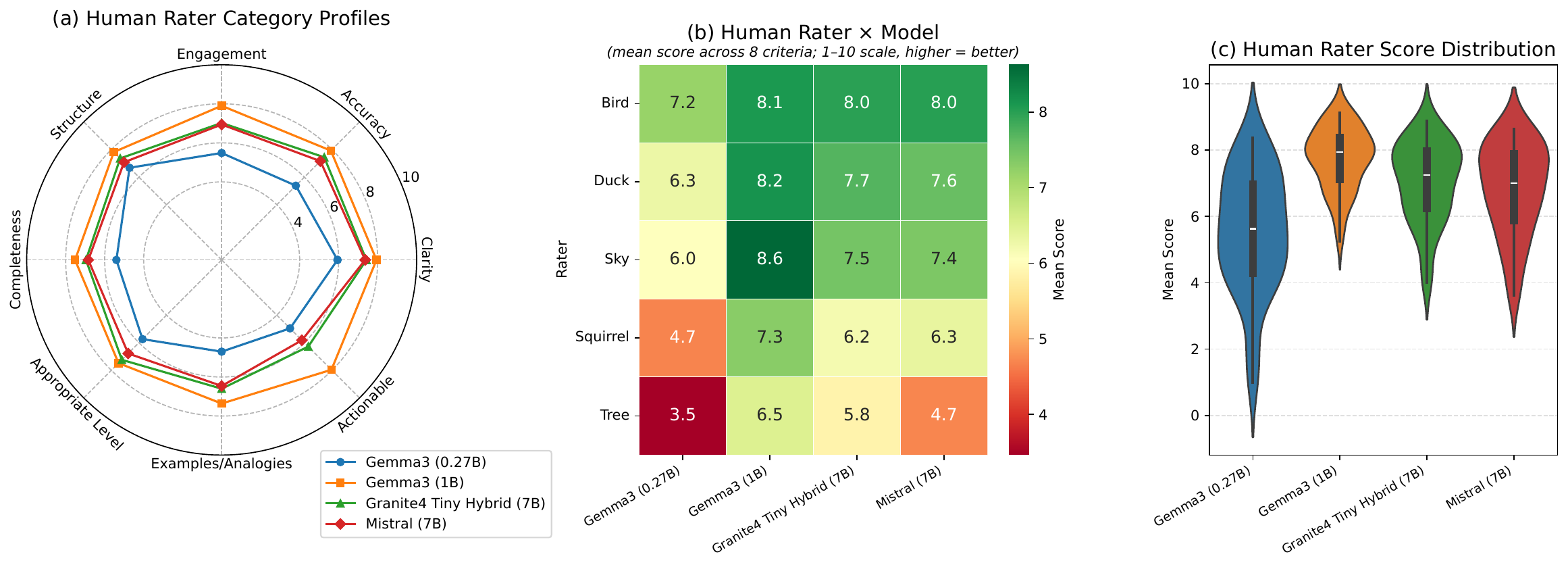}
\caption{Human annotation validation across five independent raters and four models.
(a)~Mean rater scores per model across the eight teaching criteria; markers distinguish models. 
Gemma3~0.27B scored lowest in all metrics.
(b)~Rater-by-model heatmap (mean score across 8 criteria; 1--10 scale, higher = better). 
All raters preserve the same column gradient despite differing baselines, consistent with high consistency ICC\,(C,k)\,=\,0.84 and low absolute agreement $\alpha = 0.210$.
(c)~Score distributions per model with colours matching~(a), showing wider variance for Gemma3~0.27B and tighter clustering for the three larger models.
Rater names are anonymised pseudonyms.}
\label{fig:human_validation}
\end{figure*}

Raw inter-rater agreement stood low (Krippendorff's $\alpha = 0.210$~\cite{krippendorff2011}). 
Intraclass Correlation Coefficient (ICC) analysis attributed this to scale calibration rather than ranking disagreement: single-rater consistency reached ICC(C,1)\,=\,0.51, and the five-rater average achieved ICC(C,k)\,=\,0.84 [0.75, 0.91]. 
Removing each rater's mean offset raised $\alpha$ to 0.517. Per-criterion agreement peaked at Accuracy ($\alpha = 0.277$) and Actionable ($\alpha = 0.230$), and fell lowest at Appropriate Level ($\alpha = 0.014$) and Structure ($\alpha = 0.078$). 
Figure~\ref{fig:human_validation} presents category profiles, rater-by-model scores, and score distributions across all raters.

\begin{table}[t]
\centering
\small
\renewcommand{\arraystretch}{1.3}
\caption{Human mean scores vs.\ GPT-4o-mini automated scores for the four validated models.}
\label{tab:human_vs_gpt}
\begin{tabular}{|l|c|c|c|}
\hline
\textbf{Model} & \textbf{Human} & \textbf{GPT} & \textbf{$\Delta$} \\
\hline
Gemma3 (1B)             & 7.74 & 8.5 & $-$0.76 \\
Granite4 Tiny Hybrid (7B)    & 7.04 & 7.8 & $-$0.76 \\
Mistral (7B)            & 6.81 & 8.0 & $-$1.20 \\
Gemma3 (0.27B)          & 5.54 & 5.8 & $-$0.27 \\
\hline
\end{tabular}
\end{table}

\subsection{Summary of Empirical Findings}
\label{results:summary-findings}

\begin{enumerate}
\item MMLU performance varies widely (19.0 to 57.2\%), topped by Qwen3 1.7B.
\item On RPi~4, throughput spans 0.6 to 8.7 TPS (mean 2.2 TPS), and energy efficiency spans 0.23 to 3.35 TPJ (mean 0.82 TPJ).
\item Teaching effectiveness ranges from 5.4 to 8.5, peaking at Gemma3 1B (8.5/10); 
high teaching scores do not require high MMLU (e.g., Gemma3 1B has 24.0\% MMLU).
\item Model selection indicates requiring explicit prioritisation between knowledge accuracy, responsiveness, and energy efficiency.
\item Human raters confirmed the automated teaching-effectiveness ranking (r = 0.967, n = 4 models, 5 raters), with GPT-4o-mini scoring 0.75 points higher on average.
\end{enumerate}

\section{Discussion}
\label{sec:discussion}

This section interprets benchmark results in the context of deploying language models on resource-constrained social-educational robots. 
We discuss implications of observed trade-offs for system design and model selection in the context of RSC-like robots \cite{baksh_open-source_2024}.

\subsection{Performance Trade-offs on Edge Hardware}

Results suggest that deploying LMs on RPi-class hardware entails inherent trade-offs among knowledge accuracy, responsiveness, and energy efficiency. 
Models producing higher MMLU accuracy tend to exhibit increased latency and lower throughput.
Meanwhile, more responsive models provide limited factual depth. 
These differences can be substantial: throughput and energy efficiency vary by more than an order of magnitude across evaluated models, while MMLU accuracy ranges from a random baseline up to 57\%.

Responsiveness is critical for social robots in real-time applications.  
Latency in the form of multi-second delays disrupts conversation flow, reducing perceived intelligence and rendering some high-accuracy models impractical despite strong benchmark performance. 
On the other hand, very small models offer excellent responsiveness and efficiency but may lack sufficient knowledge for sustained educational interaction. 
However, it is important to note that social-educational robots may be able to rely on ``thinking'' gestures to more naturally adjust user expectations for conversation flow when using slower models.
Overall, these findings suggest that edge deployment scenarios should evaluate LMs using multidimensional metrics rather than relying solely on accuracy.

Storage also presents a practical constraint: RPi deployments typically rely on microSD cards of 32--64\,GB, on which the operating system, inference framework, and application logic already consume a significant share. 
A multi-model architecture (Section~\ref{sec:arch_implications}) compounds this pressure: deploying Gemma3 1B and Granite4 Tiny Hybrid alongside Ollama consumes 6--7\,GB, a non-trivial fraction of a 32\,GB SD card.

\subsection{Implications for Educational Interaction}

Teaching effectiveness does not correlate monotonically with either MMLU accuracy or inference speed. 
Several models with moderate or low MMLU performance achieved strong teaching effectiveness scores, while some higher-accuracy models did not. 
This suggests that pedagogical usefulness depends on factors beyond factual recall, including explanation structure, clarity, and adaptability.

In educational settings, being able to guide learners, scaffold reasoning, and provide clear explanations may outweigh maximising factual coverage. 
However, models performing near the random MMLU baseline remain limited in robustness and risk of producing factually incorrect explanations. 

Cloud-based inference imposes recurring API costs and demands reliable internet connectivity, conditions that many schools in under-resourced or rural settings cannot meet. 
By demonstrating that educationally useful models operate on a Raspberry Pi~4 8\,GB board (recommended retail price under EUR\,100 excluding peripherals), this work points toward a deployment model where a one-time hardware purchase replaces ongoing subscription fees.  
Open-weight licensing further supports equitable access by permitting institutional inspection and redistribution.

Our human validation study (Section~\ref{results:human-validation}) corroborates these automated rankings ($r = 0.967$), lending practical support to LLM-as-judge pipelines for comparative model assessment. 
The low raw inter-rater agreement ($\alpha = 0.210$) arose from scale-calibration differences rather than substantive disagreement, as bias correction more than doubled $\alpha$ and the averaged ICC reached 0.84. 
These findings suggest that future evaluations could adopt automated rating with periodic human calibration checks, rather than requiring full human annotation. 
Rater comments further identified response completeness as the primary failure mode for sub-1B models, a finding that reinforces the quantitative scores and carries direct implications for minimum-parameter thresholds in deployment.

\subsection{Architectural Implications for the RSC (Future Work)}
\label{sec:arch_implications}

\begin{figure}[t]
\centering
\includegraphics[width=0.7\linewidth]{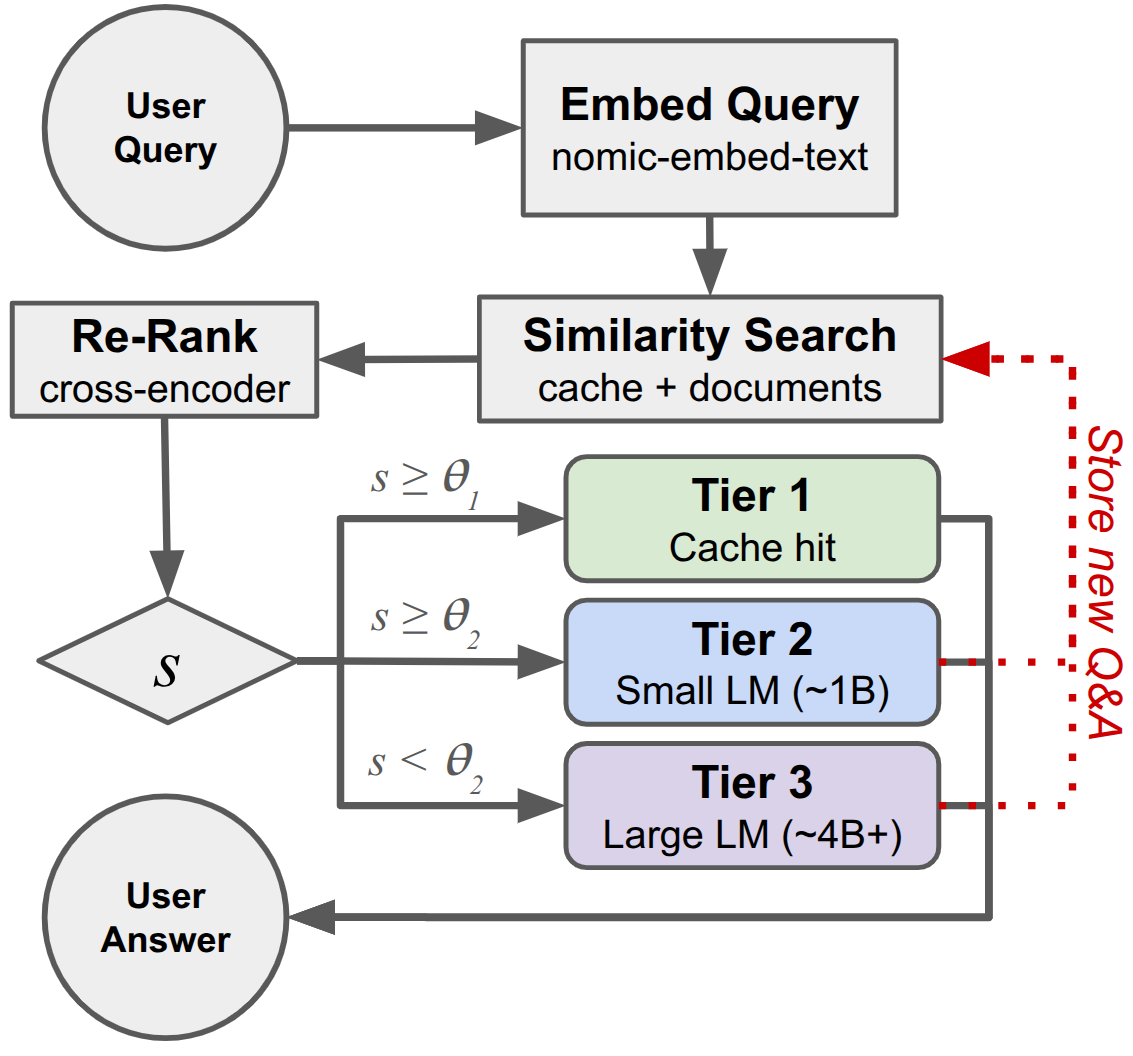}
\caption{Proposed three-tier local inference architecture for the RSC.
A sentence-embedding model (\textit{nomic-embed-text}) encodes each query and a similarity search matches it against cached Q\&A pairs and document chunks. 
A \textit{cross-encoder} re-ranks then refines the candidate scores, and the highest similarity score~$s$ routes the query to one of three tiers: a cached response ($s \geq \theta_1$), a small local LM with retrieved context ($s \geq \theta_2$), or a larger local LM ($s < \theta_2$), where $\theta_1 > \theta_2$ are configurable thresholds.
Tiers~2 and~3 write their answers back to the cache.}
\label{fig:architecture}
\end{figure}

Observed trade-off patterns motivate an architecture that avoids reliance on a single model for all interaction scenarios. 
We sketch a preliminary, as-yet-unvalidated design: a multi-tier solution for the RSC (Figure~\ref{fig:architecture}) that would dynamically balance latency and accuracy across interaction contexts.
Routing thresholds $\theta_1$ and $\theta_2$ are configurable, but their empirical calibration remains future work.

This architecture treats model capabilities as complementary rather than competitive, reserving higher-latency inference for cases where it provides a clear benefit. 
While not evaluated in a live deployment, our benchmark results quantitatively support this design and, crucially, demonstrate its feasibility on edge hardware.
In summary, these design implications should be interpreted in light of practical deployment constraints on edge devices, including sustained thermal load and model-specific inference behaviour.

\subsection{Limitations}
\label{limitations}

Several constraints warrant consideration. 
Inference performance on RPi devices is sensitive to sustained thermal load, and all hardware metrics derive from single runs without variance estimates, limiting statistical confidence in absolute values while preserving relative rankings.
Furthermore, the linear CPU-load-to-current model underlying our TPJ estimate was calibrated on the RPi~4 and applied unchanged on the RPi~5, whose higher full-load current envelope likely inflates absolute Pi~5 TPJ values; relative rankings across models on either platform remain informative.
For reasoning-oriented LMs, latency and throughput metrics reflect visible token emission rather than total internal computation, potentially yielding optimistic TTFT and TPS values.
Additionally, we derive TPS from streamed output chunks rather than exact tokeniser counts, which may introduce minor discrepancies across models and runtimes; however, this does not affect relative comparisons within the benchmark. 
LLM-as-judge evaluation carries known biases along dimensions including response length, style, position, and self-similarity to the evaluator. 
Our human validation study confirmed rank-order agreement with the automated rater but revealed GPT-4o-mini's systematic positive bias of 0.75~points on the 1--10 scale and low raw inter-rater reliability. 
Together, findings suggest absolute GPT-4o-mini scores should be treated as indicative rather than calibrated, and that subjective criteria would benefit from more detailed rubric anchoring in future work.

Finally, our six-category MMLU subset ensures tractability but may not fully capture broader domain knowledge. 
These limitations do not affect relative trends across models but constrain the generalisability of absolute values beyond the evaluated configuration.

\section{Conclusion}

This paper presents a reproducible benchmark of 25 open-source language models for deployment on resource-constrained social-educational robots, evaluating inference efficiency, knowledge performance, and teaching effectiveness.
No single model optimises all dimensions, and teaching effectiveness does not strictly correlate with MMLU accuracy or throughput.
These findings motivate a multi-tier local inference architecture, combining caching, lightweight local models, and higher-capacity inference on demand, and provide practical guidance for privacy-preserving social robotics.

\subsection*{CRediT Statement} 
D. Lamouille: Conceptualisation, Software, Investigation, Methodology, Data curation, Writing - original draft. 
M.B. Zorec: Conceptualisation, Data curation, Investigation, Software, Supervision, Methodology, Visualisation, Writing - original draft, review \& editing. 
F. Baksh: Conceptualisation, Supervision, Writing - review \& editing. 
K. Kruusam\"ae: Funding acquisition, Writing - review \& editing.

\bibliographystyle{IEEEtran}
\bibliography{references}

\end{document}